\def\year{2020}\relax
\documentclass[letterpaper]{article} 
\usepackage{aaai20}  
\usepackage{times}  
\usepackage{helvet} 
\usepackage{courier}  
\usepackage[hyphens]{url}  
\usepackage{graphicx} 
\usepackage{graphicx}  
\usepackage{listings}
\usepackage[ruled,linesnumbered]{algorithm2e}
\usepackage[hyphens]{url}
\usepackage{multirow}
\usepackage{subcaption}
\usepackage{amsmath,amssymb,amsthm}
\usepackage{bbm}
\usepackage{multirow}
\usepackage{adjustbox}
\usepackage{tikz}
\usetikzlibrary{positioning,arrows,shapes}
\usepackage{subfiles}
\lstdefinelanguage{PDDL}
{
  sensitive=false,    
  morecomment=[l]{;}, 
  alsoletter={:,-},   
  morekeywords={
    define,domain,problem,not,and,or,when,forall,exists,either,
    :domain,:requirements,:types,:objects,:constants,
    :predicates,:action,:parameters,:precondition,:effect,
    :fluents,:primary-effect,:side-effect,:init,:goal,
    :strips,:adl,:equality,:typing,:conditional-effects,
    :negative-preconditions,:disjunctive-preconditions,
    :existential-preconditions,:universal-preconditions,:quantified-preconditions,
    :functions,assign,increase,decrease,scale-up,scale-down,
    :metric,minimize,maximize,
    :durative-actions,:duration-inequalities,:continuous-effects,
    :durative-action,:duration,:condition
  }
}

\title{TempAMLSI : Temporal Action Model Learning based on Grammar Induction}
\author{
Maxence Grand, Damien Pellier \and Humbert Fiorino\\
Laboratoire d'Informatique de Grenoble - MARVIN\\
Université Grenoble Alpes, Saint Martin d'Hères, France\\
\{Maxence.Grand, Damien.Pellier, Humbert.Fiorino\}@univ-grenoble-alpes.fr
}

\begin{document}
\maketitle

\begin{abstract}
Hand-encoding PDDL domains is generally accepted as difficult, tedious and error-prone. The difficulty is even greater when temporal domains have to be encoded. Indeed, actions have a duration and their effects are not instantaneous. In this paper, we present TempAMLSI, an algorithm based on the AMLSI approach able to learn temporal domains. TempAMLSI is based on the classical assumption done in temporal planning that it is possible to convert a non-temporal domain into a temporal domain. TempAMLSI is the first approach able to learn temporal domain with single hard envelope and Cushing's intervals. We show experimentally that TempAMLSI is able to learn accurate temporal domains, i.e., temporal domain that can be used directly to solve new planning problem, with different forms of action concurrency.
\end{abstract}

\section{Introduction}

Thanks to description languages like PDDL \cite{pddl}, AI planning has become more and more important in many application fields. One reason is the versatility of PDDL  to represent \textit{durative actions} \cite{TemporalPDDL}, i.e. actions that have a duration, and whose preconditions and effects must be satisfied and applied at different times.

Temporal PDDL domains have different levels of required action concurrency \cite{cushing}. Some of them are {\em sequential}, which means that all the plan parts containing overlapping durative actions can be rescheduled into a completely sequential succession of durative actions: each durative action starts after the previous durative action is terminated. One important property of sequential temporal domains is that they can be rewritten as classical domains, and therefore used by classical non-temporal planners. Some temporal domains require different forms of action concurrence such as \textit{Single Hard Envelope} (SHE) \cite{env}. SHE is a form of action concurrency where a durative action can be executed only if another durative action called the {\em envelope}, is executed simultaneously. This is due to the fact that the enveloped durative action needs a resource, during all its execution, added at the start of the execution of the envelope and deleted at the end of the execution of the envelope. One important property of SHE temporal domains is that they cannot be sequentially rescheduled.

Hand-encoding PDDL domains is generally considered difficult, tedious and error-prone by experts, and this is even more harder with action concurrency. It is therefore essential to develop tools allowing to acquire temporal domains.

To facilitate PDDL domain acquisition, different machine learning algorithms have been proposed. First, for classical domains as for instance, ARMS \cite{arms}, SLAF \cite{slaf}, Louga \cite{louga}, LSONIO \cite{lsonio}, LOCM \cite{locm}, IRale \cite{irale}, PlanMilner \cite{planmilner}. In these approaches, training data are either (possibly noisy and partial) intermediate states and plans previously generated by a planner, or randomly generated action sequences. These learning techniques are promising, but they cannot be used to learn temporal domains. \cite{pddlCSP} have proposed an algorithm to learn temporal domains using CSP techniques, however their approach is limited to sequential temporal domains. To our best knowledge, there is no learning approach for both SHE and sequential temporal domains.


Several temporal planners \cite{lpgp,crikey,tp,stp} attempt to exploit classical planning techniques for temporal planning. These planners convert a temporal domain into a classical domain, i.e. a domain containing non-durative action, generate a plan using this classical domain, and use rescheduling techniques to make the plan compatible with durative actions. Our contribution is to propose an approach exploiting the conversion of temporal domains into classical domains, initially proposed to solve temporal planning problems, for the temporal domain learning task. More precisely, in this work we assume that is possible to reduce the temporal domain learning task into the classical domain learning task.

In this paper, we present TempAMLSI, a learning algorithm for temporal domains including different levels of required action concurrency. TempAMLSI is built on AMLSI \cite{amlsi_keps}, a PDDL domain learner based on grammar induction. Like AMLSI, TempAMLSI takes as input feasible and infeasible action sequences to frame what is allowed by the targeted domain. More precisely, TempAMLSI consists of three steps: (1) TempAMLSI converts temporal sequences into non-temporal sequences, (2) TempAMLSI learns a classical domain containing non-durative action using AMLSI, and (3) converts it into a temporal domain containing durative actions.


The rest of the paper is organized as follows. In section \ref{sec:problem} we present a problem statement. In section \ref{sec:backgroung_amlsi} we give some backgrounds on AMLSI approach and, in section \ref{sec:method}, we detail TempAMLSI steps. Finally, section \ref{sec:experiments} evaluates the performance of TempAMLSI on IPC temporal benchmarks.

\section{Problem Statement}\label{sec:problem}

This section introduces a formalization of planning domain learning which consisting in learning a transition function of a grounded planning domain, and in expressing it as PDDL operators.

In classical planning, world states $s$ are modeled as sets of logical propositions, and actions change the world states. Formally, let $S$ be the set of all the propositions modeling properties of world, and $A$ the set of all the possible actions in this world. A {\it state} $s$ is a subset of $S$ and each action $a \in A$ is a triple of proposition sets $(\rho_a, \epsilon^{+}_a, \epsilon^{-}_a)$, where $\rho_a, \epsilon^{+}_a, \epsilon^{-}_a \subseteq S$, and $\epsilon^{+}_a \cap \epsilon^{-}_a = \emptyset$. $\rho_a$ are the preconditions of action $a$, that is, the propositions that must be in the state before the execution of action $a$. $\epsilon^{+}_a$ and $\epsilon^{-}_a$ are respectively the positive (add list) and the negative (del list) effects of action $a$, that is, the propositions that must be added or deleted in $s$ after the execution of the action $a$. Therefore, learning a classical planning domain consists in learning the deterministic state transition function $\gamma : S \times A \rightarrow S$ defined as: $\gamma(s, a) = (s \cup \epsilon^{+}_a) \setminus \epsilon^{-}_a$ where
$\gamma(s,a)$ exists if $a$ is applicable in $s$, i.e., if and only if $\rho_a \subseteq s$.\\


In temporal planning~\cite{TemporalPDDL}, states are defined as in classical planning. However, the action set $A$ is a set of durative actions. A {\it durative action} $a$ is composed of:
\begin{itemize}
    \item $d_a$; the duration
    \item $\rho_a(s), \rho_a(e), \rho_a(o)$: preconditions of $a$ at start, over all, and at end, respectively.
    \item $\epsilon^{+}_a(s), \epsilon^{+}_a(e)$: positive effects of $a$ at start and at end, respectively.
    \item $\epsilon^{-}_a(s), \epsilon^{-}_a(e)$: negative effects of $a$ at start and at end, respectively.
\end{itemize}

The semantics of durative actions is defined in terms of two discrete events $start_a$ and $end_a$, each of which is naturally expressed as a classical action. Starting a durative action $a$ in state $s$ is equivalent to applying the classical action $start_a$ in $s$, first verifying that $\rho_{start_a}$ holds in $s$. Ending $a$ in state $s'$ is equivalent to applying $end_a$ in $s'$, first by verifying that $\rho_{end_a}$ holds in $s'$. $start_a$ and $end_a$ are defined as follows:
\begin{equation*}
    \resizebox{1\hsize}{!}{$
    \begin{array}{ccc}
        start_a : \rho_a(s) = \rho_{start_a} & \epsilon^{+}_a(s) = \epsilon^{+}_{start_a} & \epsilon^{-}_a(s) = \epsilon^{-}_{start_a}\\
        end_a :\rho_a(e) = \rho_{end_a} & \epsilon^{+}_a(e) = \epsilon^{+}_{end_a} & \epsilon^{-}_a(e) = \epsilon^{-}_{end_a}\\
    \end{array}
    $}
\end{equation*}

\begin{figure}[t]
    \centering
    \includegraphics{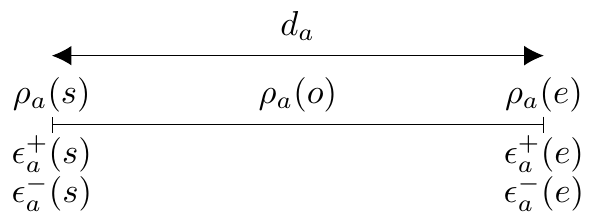}
    \caption{Structure of a durative action $a$}
    \label{fig:structure_durative}
\end{figure}

This process is restricted by the duration of $a$, denoted $d_a$ and the over all precondition. Event $end_a$ has to occur exactly $d_a$ time units after $start_a$ and the over all precondition has to hold in all states between $start_a$ and $end_a$. Although $a$ has a duration, its effects apply instantaneously at the start and end of $a$, respectively. The preconditions $\rho_a(s)$ and $\rho_a(e)$ are also checked instantaneously, but $\rho_a(o)$ has to hold for the entire duration of $a$. The structure of a durative action is summarized in the Figure~\ref{fig:structure_durative}.

A {\it temporal action sequence} is a set of action-time pairs $\pi = \{(a_1 , t_1 ), \ldots , (a_n , t_n )\}$. Each action-time pair $(a, t) \in \pi$ is composed of a durative action $a \in A$ and a scheduled start time $t$ of $a$, and induces two events $start_a$ and $end_a$ with associated timestamps $t$ and $t+d_a$, respectively. Events $start_a$ (resp. $end_a$) is applied in the state $s_t$ (resp. $s_{t+d_a}$), $s_t$ (resp. $s_{t+d_a}$) being a state time-stamped with $t$ (resp. $t+d_a$). Then, the temporal transition function $\gamma$ to learn can be rewritten as: $\gamma(s,a,t) = (\gamma(s_t, start_a), \gamma(s_{t+d_a}, end_a))$. The transition function $\gamma(s,a,t)$ is defined if and only if: $\rho_a(s) \in s_t$, $\rho_a(e) \in s_{t+d_a}$ and $\forall t'$ such that $t \leq t' \leq t+d_a$ $\rho_a(o) \in s_{t'}$.\\

\begin{figure}[t]
    \centering
    \includegraphics[width=0.5\linewidth]{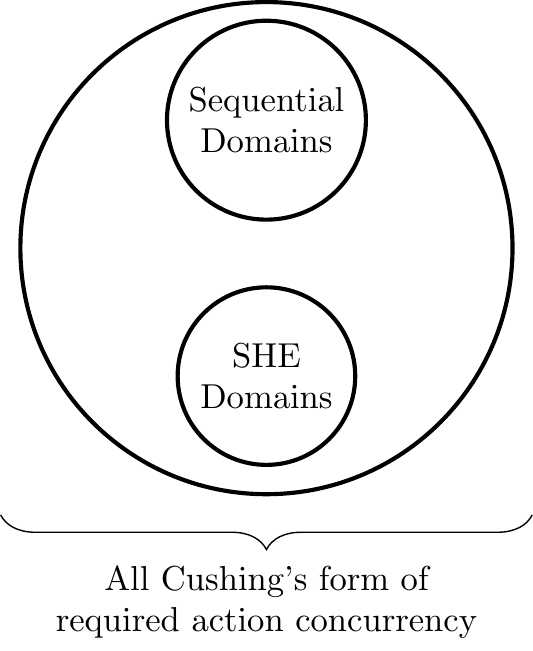}
    \caption{Different form of required action concurrency}
    \label{fig:form_concurrency}
\end{figure}

To learn the state transition function $\gamma$ and to express it as a PDDL temporal domain, we assume that:
\begin{itemize}
\item we are able to observe temporal sequences of state/action defined recursively as follows:
\begin{equation*}
    \resizebox{1\hsize}{!}{$
    \begin{array}{l}
    \Gamma(s,\pi) =
    \begin{cases}
    [s]& \text{if } \pi = \emptyset\\
    [s]& \text{if } \gamma(s, a_0, t_0) \text{ undef}\\
    [s]+\Gamma(\gamma(s, a_0, t_0), [(a_1, t_1), .. , (a_n, t_n)])& \text{otherwise}
    \end{cases}
    \\
    \end{array}
    $}
\end{equation*}
where observed states $s$ can be possibly partial. A partial state is a state where some propositions are missing.
\item for all action $a = (\eta_a$, $d_a$, $\rho_a(s)$, $\rho_a(e)$, $\rho_a(o)$, $\epsilon^{+}_a(s)$, $\epsilon^{+}_a(e)$, $\epsilon^{-}_a(s)$, $\epsilon^{-}_a(e))$ in the sequences of state/action, $\eta_a$, the name of $a$ is known, $d_a$ is a known constant, and $\rho_a(s)$, $\rho_a(e)$, $\rho_a(o)$, $\epsilon^{+}_a(s)$, $\epsilon^{+}_a(e)$, $\epsilon^{-}_a(s)$, and $\epsilon^{-}_a(e)$ are unknown.
\item learned temporal domains can required different forms of action concurrency (see Figure \-- \ref{fig:form_concurrency}) such as {\it Single Hard Envelope} (SHE) \cite{env}. SHE is a form of action concurrency where the execution of a durative action $a$
is required for the execution of a second durative action $a'$. Formally, a SHE is a durative action $a'$ that adds a proposition $p$ at start and deletes it at end while $p$ is an over all precondition of a durative action $a$. Contrary to sequential temporal domains, for temporal domains containing SHE there exists temporal action sequences that cannot be sequentially rescheduled. For instance, the Match domain and the the following actions:
\begin{itemize}
    \item $MEND(f~m)$ such that $(light~m) \in \rho_{MEND(f~m)}(o)$
    \item  $LIGHT(m)$ such that $(light~m) \in \epsilon^{+}_{LIGHT(m)}(s)$ and $(light~m) \in \epsilon^{-}_{LIGHT(m)}(e)$
\end{itemize}
The durative action $MEND(f~m)$ cannot start before the start of the durative action $LIGHT(m)$ and $MEND(f~m)$ cannot end after the end of $LIGHT(m)$, so $MEND(f~m)$ have to start after the start of $LIGHT(m)$ and $MEND(f~m)$ have to end before the end of $LIGHT(m)$, it is therefore impossible to sequentially reschedule any temporal action sequences containing these actions. Finally, note that there is other forms of required action concurrency than SHE \cite{cushing}.

\end{itemize}

\section{Background on AMLSI}\label{sec:backgroung_amlsi}

\begin{figure}[t]
    \centering
    \includegraphics[scale=0.9]{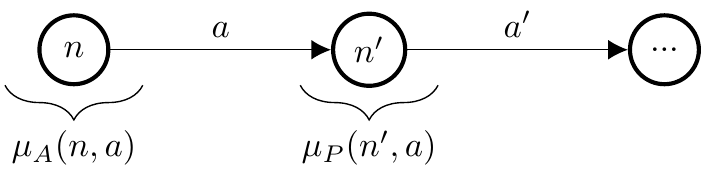}
    \caption{Mapping overview}
    \label{fig:mapping}
\end{figure}

AMLSI takes as inputs two training datasets, $I_{+}$ and $I_{-}$, and outputs a PDDL domain. $I_{+}$ (positive samples) contains the observed feasible state/action sequences, and $I_{-}$ (negative samples) contains infeasible state/action sequences computed from $I_{+}$. $I_{+}$ is generated by using random walks. When positive samples are built, each first infeasible action induces a new negative sample (where only the last action is infeasible) recorded in $I_{-}$.

The AMLSI algorithm consists of 3 steps: (1) AMLSI learns a Deterministic Finite Automaton (DFA) corresponding to the regular grammar generating the action sequences in $I_{+}$ and forbidding those in $I_{-}$; (2) AMLSI induces the PDDL operators from the learned DFA; (3) finally, AMLSI refines these operators to deal with noisy and partial state observations.

The first step consisting in learning the DFA is carried out by using a variant of the classic algorithm for learning regular grammar RPNI \cite{rpni}. The RPNI algorithm used by AMLSI deals with PDDL features and encodes the links between preconditions and effects of actions to speedup the learning process (see \cite{amlsi_ictai} for a complete description). Formally, the learned DFA is a quintuple $(A, N, n_0, \gamma, F)$, where $A$ is the set of actions, $N$ is the set of nodes, $n_0 \in N$ is the initial node, $\gamma$ is the node transition function, and $F \subseteq N$ is the set of final nodes.

The second step consists in generating the PDDL operators of the planning domain to learn. To carry out this step, AMLSI must first know which node of the DFA corresponds to which observed state. Thus, AMLSI maps the pairs "node, action" in the DFA with the pairs "state, action" of all $\pi \in I_{+}$ (see Figure \ref{fig:mapping}) and labels the propositions of the node that represents the preconditions and the effects of the action transition in the DFA. Therefore, there are two different labels for a node: the (A)nte label $\mu_A$ and (P)ost label $\mu_P$. $\mu_A(n,a)$ (resp. $\mu_P(n,a)$) gives the intersection of state set before (resp. after) the execution of the transition $a$ in node $n$: $a$ is an outcoming (resp. incoming) edge of $n$ in the DFA. Once the labels are computed, AMLSI induces the preconditions and effects of the planning operators. The preconditions of an operator $o$ are the set intersection of all the labels $\mu_A(n, a)$ such that $a$ is an instance of $o$ and $a$ is an outgoing transitions of the node n. Formally, $p \in \rho_o$ if and only if for all $a$ instance of $o$,
\begin{equation*}\label{eq:generation_precondition}
    p \in \mu_A(n, a)
\end{equation*}
Then, the negative effects $\epsilon_o^{-}$ of an operator $o$ are computed as the set intersection of the propositions present before the execution of all the actions $a$ instances of $o$, and never after. Formally, $p \in \epsilon_o^{-}$ if and only if for all $a$ instance of $o$:
\begin{equation*}\label{eq:generation_positive_effect}
    p \in \mu_A(n, a) \wedge p \not\in \mu_P(n,a)
\end{equation*}
Finally, the positive effects $\epsilon_o^{+}$ of an operator $o$ are computed similarly: $p \in \epsilon_o^{+}$ if and only if for all $a$ instance of $o$:
\begin{equation*}\label{eq:generation_negative_effect}
    p \not\in \mu_A(n, a) \wedge p \in \mu_P(n,a)
\end{equation*}

The last step consists in refining the PDDL operators induced at step 2 to deal with noisy and partial state observations. First of all, AMLSI starts by refining the operator effects to ensure that the generated operators allow to regenerate the induced DFA. To that end, AMLSI adds all effects allowing to ensure that each transition in the automaton are feasible. Then, AMLSI refines the preconditions of the operators. AMLSI makes the following assumptions as in \cite{arms}: The negative effects of an operator must be used in its preconditions. Thus, for each negative effect of an operator, AMLSI adds the corresponding proposition in the preconditions. Since effect refinements depend on preconditions and precondition refinements depend on effects, AMLSI repeats these two refinements steps until convergence, i.e., no more precondition or effect is added. Finally, AMLSI performs a Tabu Search to improve the PDDL operators independently of the induced DFA, on which operator generation is based. The fitness score used to evaluate a candidate set $D$ of PDDL operators is:

\begin{equation*}\label{eq:tabu}
\begin{array}{ll}
J(D | I_{+}, I_{-}) = & J_\rho(D | I_{+}) + J_\epsilon(D | I_{+}) +\\
&J^{+}(D | I_{+}) + J^{-}(D | I_{-})\\
\end{array}
\end{equation*}
where :
\begin{itemize}
\item \scalebox{0.9}[1]{$J_\rho(D | I_{+}) = \sum\limits_{\pi \in I_{+}}
\sum\limits_{(s,a) \in \Gamma(s_0, \pi)} Accept(\rho_a, s) - Reject(\rho_a, s)$} computes the
fitness score for the preconditions of the actions $a$. $Accept(\rho_a, s)$ counts the number of
positive and negative preconditions in the observed state $s$, and
$Reject(\rho_a, s)$ counts the number of positive and negative preconditions that are not in $s$.
\item \scalebox{0.7}[1]{$J_\epsilon(D | I_{+}) = \sum\limits_{\pi \in I_{+}} \sum\limits_{(s,a,s') \in \Gamma(s_0, \pi)} Equal(s', \gamma(s,a)) - Different(s', \gamma(s,a))$}
computes the fitness score for the effects of the action $a$. $s$ (resp. $s'$) is the observed states before (resp. after) the execution of the action $a$. $Equal(s', \gamma(s,a))$ counts the number of similar propositions in $s'$ and $\gamma(s,a)$, and $Different(s', \gamma(s,a))$ counts the differences.
\item \scalebox{0.85}[1]{$J^{+}(D | I_{+}) = \sum\limits_{\pi \in I_{+}}
|\pi| \times \mathbbm{1}_{Accept(D, \pi)}$ where $\mathbbm{1}_{Accept(D, \pi)} = 1$}
if and only if $D$ can generate the positive sample $\pi$. $|\pi|$ is the length of $\pi$. $J^{+}(D | I_{+})$ is weighted by the length of $\pi \in I_{+}$ because $I_{+}$ is smaller than $I_{-}$
\item \scalebox{0.95}[1]{$J^{-}(D | I_{-}) = \sum\limits_{\pi \in I_{-}}
\mathbbm{1}_{Reject(D, \pi)}$} where $\mathbbm{1}_{Reject(D, \pi)} = 1$
if and only if $D$ cannot generate the negative samples $\pi$.
\end{itemize}

Once the Tabu Search reaches a local optimum, AMLSI repeats all the three refinement steps until convergence.

\section{Temporal AMLSI}\label{sec:method}

\begin{figure*}[!t]
    \centering
    \includegraphics[width=0.95\linewidth]{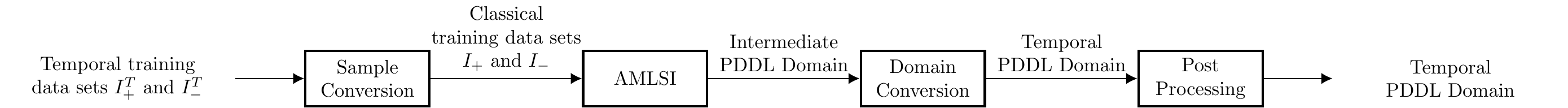}
    \caption{Overview of the TempAMLSI approach}
    \label{fig:approach}
\end{figure*}

\begin{figure}[!t]
    \centering
    \begin{subfigure}{\hsize}
        \begin{adjustbox}{width=1\textwidth}
        \begin{lstlisting}[language=PDDL]
(:action MEND-START
 :parameters (?f - fuse ?m - match)
 :precondition (and (handfree) (light ?ma))
 :effect (and (not (handfree))))

(:action MEND-END
 :parameters (?f - f ?m - m)
 :precondition (and (light ?m))
 :effect (and (mended ?f) (handfree)))
        \end{lstlisting}
        \end{adjustbox}
        \caption{Classical declaration of the operator MEND}
        \label{algo:classical}
    \end{subfigure}

    \begin{subfigure}{\hsize}
        \begin{adjustbox}{width=1\textwidth}
        \begin{lstlisting}[language=PDDL]
(:durative-action MEND
 :parameters (?f - fuse ?m - match)
 :duration (= ?duration 2)
 :condition (and (at start (handfree))
    (over all (light ?m)))
 :effect (and (at start (not (handfree)))
     (at end (mended ?f)) (at end (handfree))))
        \end{lstlisting}
        \end{adjustbox}
        \caption{Durative declaration of the operator MEND}
        \label{algo:temporal}
    \end{subfigure}
    \caption{Comparison between the durative declaration and the classical declaration of the operator MEND of the Match domain.}
    \label{fig:domain_conversion}
\end{figure}

The TempAMLSI approach is summarized by the Figure \-- \ref{fig:approach}. After having built the samples containing temporal sequences (including both feasible and infeasible sequences), TempAMLSI converts the temporal samples into non-temporal sequences (see Section \-- \ref{sec:method:sample}). We use the AMLSI algorithm to learn an intermediate classical PDDL domain, and then convert it into a temporal PDDL domain (see Section \-- \ref{sec:method:domain}). Finally TempAMLSI performs a temporal refinement step (see Section \-- \ref{sec:method:TR}). This step allows to take into account the temporal constraints which are not present in the non-temporal sequences.

\subsection{Sample conversion}\label{sec:method:sample}

Let us take a sample containing $\pi^T$, $\pi^T$ being a positive temporal sequence such that:
\begin{equation*}\label{eq:ex:temporalsequence}
    \resizebox{1\hsize}{!}{$
    \begin{array}{ll}
        \pi^T= &\{(0,LIGHT(m)), (0.5,MEND(f_1,m)), (2.6,MEND(f_2,m))\}\\
    \end{array}
    $}
\end{equation*}

We can convert each durative action in the following way: Each durative action $a$ is converted into two event actions $start(a)$ and $end(a)$. After conversion, we have the following sample:

\begin{equation*}\label{eq:ex:classicalSequence}
    \resizebox{1\hsize}{!}{$
    \begin{array}{ll}
        \pi= &\{start(LIGHT(m)), start(MEND(f_1,m)), end(MEND(f_1,m)),\\
                 &start(MEND(f_2,m)), end(MEND(f_2,m)), end(LIGHT(m))\}\\
    \end{array}
    $}
\end{equation*}
 Negative temporal sequences are converted in the same way.

\subsection{Domain conversion}\label{sec:method:domain}

After having learned the classical PDDL domain with AMLSI, TempAMLSI converts it PDDL operators into a temporal one in the following way:
\begin{equation*}\label{eq:domainConversion}
    \resizebox{1\hsize}{!}{$
    \begin{array}{l}
        \rho_a(s)=\rho_{start(a)}\setminus\rho_{end(a)},~\epsilon^{+}_a(s)=\epsilon^{+}_{start(a)},~\epsilon^{-}_a(s)=\epsilon^{-}_{start(a)}\\
        \rho_a(e)=\rho_{end(a)}\setminus\rho_{start(a)},~\epsilon^{+}_e(s)=\epsilon^{+}_{end(a)},~\epsilon^{-}_e(s)=\epsilon^{-}_{end(a)}\\
        \rho_a(o)=\rho_{start(a)}\cap\rho_{end(a)}\\
    \end{array}
    $}
\end{equation*}

First of all, At start (resp. at end) effects are the effects of start (resp. end) classical operators. Then, Overall preconditions are the intersection of preconditions of start and end classical operators. Finally, At start (resp. at end) preconditions are preconditions of the start (resp. end) classical operator excluding end (resp. start) preconditions. The Figure \-- \ref{fig:domain_conversion} gives an example of action conversion for the MEND operator of the Match domain.

\subsection{temporal refinement}\label{sec:method:TR}

For domains that are not sequential (see Section \-- \ref{sec:problem}) it is possible that some non-temporal actions sequences are both feasible and infeasible, since the feasibility of an action sequence depends on temporal constraints which are not present in non-temporal sequences. For instance, the action sequence:
\begin{equation*}
    \begin{array}{l}
        \pi_1=\{start(LIGHT(m)), start(MEND(f_1,m)\}\\
    \end{array}
\end{equation*}
could be both a negative example and the prefix of a positive example. Indeed, $\pi_1$ is the non-temporal action sequence returned after the conversion of the following negative temporal action sequence:
 \begin{equation*}
     \begin{array}{l}
         \pi^T_1=\{(0,LIGHT(m)), (4, MEND(f_1,m))\}\\
     \end{array}
 \end{equation*}
However, $\pi_1$ is also the prefix of the action sequence
 \begin{equation*}
     \begin{array}{ll}
         \pi_2=&\{start(LIGHT(m)), start(MEND(f_1,m)\\
               & end(MEND(f_1,m)), end(LIGHT(m))\}\\
     \end{array}
 \end{equation*}
which is the sequence returned by the conversion of the following positive example:
\begin{equation*}
    \begin{array}{l}
        \pi^T_1=\{(0,LIGHT(m)), (2, MEND(f_1,m))\}\\
    \end{array}
\end{equation*}

For this type of domain, it is therefore impossible for the Tabu search of the AMLSI algorithm to reach a global maximum (see Section \-- \ref{sec:backgroung_amlsi}). Let us take the equation \-- \ref{eq:tabu} and suppose that $\pi_2 \in I_{+}$ and $\pi_1 \in I_{-}$, if $J^{+}(D|I_{+})$ is maximized then $J^{+}(D|I_{+})$ cannot be maximized and vice versa. Indeed, a classical domain cannot both accept $\pi_2$ and reject $\pi_1$. However, it is necessary that the temporal domain accepts $\pi^T_2$  and rejects $\pi^T_2$ to ensure that $(light~m) \in \rho(o)_{MEND}$. The objective of the temporal refinement step is to take into account the temporal constraints which are not present in the non-temporal sequences..

The temporal refinement step is a Tabu search. The fitness score for this Tabu search is defined as follows:
\begin{equation*}\label{eq:tabuTR}
    J^T(D|I_{+}^T, I_{-}^T) = \frac{J^T_ {+}(D|I_{+}^T)+J^T_ {-}(D|I_{-}^T)}{2}
\end{equation*}
where
\begin{equation*}
    J^T_ {+}(D|I_{+}^T) = \frac{\sum\limits_{\pi^T \in I_{+}^T}\mathbbm{1}_{Accept(D, \pi^T)}}{|I_{+}^T|}
\end{equation*}
\begin{equation*}
    J^T_ {-}(D|I_{-}^T) = \frac{\sum\limits_{\pi^T \in I_{-}^T}\mathbbm{1}_{Reject(D, \pi^T)}}{|I_{-}^T|}
\end{equation*}

Then the neighborhood of a candidate domain $D$ is the set of domains where a precondition or an effect is added or removed in an operator of $D$. And the search space of the Tabu Search is the set of all possible domains compatible with the following syntax constraints:
\begin{itemize}
    \item $\rho_a(o) \cap \{\epsilon{+}_a(s) \cup \epsilon{+}_a(e)\} = \emptyset$: At start and at end positive effect cannot add propositions present in the the over all precondition.
    \item $\rho_a(s) \cap \epsilon{+}_a(s) = \emptyset$: At start positive effect cannot add propositions present in the at start precondition.
    \item $\rho_a(e) \cap \epsilon{+}_a(e) = \emptyset$: At end positive effect cannot add propositions present in the at end precondition.
    \item $\epsilon{-}_a(s) \cap \epsilon{+}_a(s) = \emptyset$: An at start effect cannot be both positive and negative.
    \item $\epsilon{-}_a(e) \cap \epsilon{+}_a(e) = \emptyset$:An at end effect cannot be both positive and negative.
    \item $\epsilon{-}_{a}(s) \cap \{\rho_{a}(o) \cup \rho_{a}(e)\} = \emptyset$: An at start effect cannot delete a proposition present in the at end and over all precondition.
    \item $\epsilon{+}_{a}(s) \cap \epsilon{+}_{a}(e) = \emptyset$: An at start effect and an at end effect cannot add the same proposition.
    \item $\epsilon{-}_{a}(s) \cap \epsilon{-}_{a}(e) = \emptyset$: An at start effect and an at end effect cannot delete the same proposition.
\end{itemize}
These syntactical constraints are base on the syntactical constraints proposed by \cite{arms} and modified to take into account the time labels of the preconditions and effects.

\section{Experiments and evaluations}\label{sec:experiments}

\begin{table}[!t]
\centering
\begin{tabular}{|c||c|c|c|c|}
\hline
Domain & \# Operators & \# Predicates  & Class\\
\hline\hline
Peg Solitaire & $1$ & $3$ & Sequential\\ \hline
Sokoban & $2$ & $3$ & Sequential \\ \hline
Parking & $4$ & $5$ & Sequential \\ \hline
Zenotravel & $5$ & $4$ & Sequential \\ \hline
Turn and Open & $5$ & $8$ & SHE \\ \hline
Match & $2$ & $4$ & SHE \\ \hline
Cushing & $3$ & $7$ & Cushing \\ \hline
\end{tabular}
\caption{\label{tab:benchmarks} Benchmark domain characteristics}
\end{table}

\subsection{Experimental setup}
Our experiments are based on 7 temporal domains (see Table \-- \ref{tab:benchmarks}). More precisely we test TempAMLSI with four Sequential domains (Peg Solitaire, Sokoban, Parking, Zenotravel), two SHE domains (Match, Turn and Open), and one domain with forms of required action concurrency defined in \cite{cushing}, this domain, called Cushing, has been proposed by \cite{stp}.

We deliberately choose the size of the test sets larger than the training sets to show TempAMLSI ability to learn accurate domains with small datasets. The training and test sets are generated as follows: at a given state $s$, we randomly choose a durative action $a$ in $A$. If $a$ is feasible, the current state is observed, and we add $a$ to the current $\pi$. This random walk is iterated until $\pi$ reaches an arbitrary length (randomly chosen between $5$ and $15$), and added to $I_{+}$. If $a$ is infeasible in the current state, the concatenation of $\pi$ and $a$ is added to $I_{-}$. In the test sets $E_{+}$ and $E_{-}$, we generate action sequences with a length randomly chosen between $1$ and $30$.

We test each domains with three different initial states over five runs, and we use five seeds randomly generated for each run. All tests were performed on an Ubuntu 14.04 server with a multi-core Intel Xeon CPU E5-2630 clocked at $2.30$ GHz with 16GB of memory.

\subsection{Evaluation Metrics}\label{sec:experiments:metrics}
\label{subsec:Metrics}
We evaluate TempAMLSI with four different metrics. The first metric, the syntactical error, is the most used metric in the literature. The other metrics, the FScore, the accuracy and the IPC score are more specific to our approach.

\begin{itemize}
\item {\em Syntactical Error :} The syntactical error $error(a)$ for an action $a$ is defined as the number of extra or missing predicates in the preconditions $\rho_a(s,e,o)$, the positive effects $\epsilon_a^{+}(s,e)$ and the negative effects $\epsilon_a^{-}(s,e)$ divided by the total number of possible predicates \cite{zhuo10}. The syntactical error for a domain with a set of actions $A$ is: $E_{\sigma} = \frac{1}{|A|} \sum_{a \in A} error(a)$.\\

\item {\em FScore:} This metric is initially used for pattern recognition and binary classification \cite{fscore}. Nevertheless, it can be used to evaluate the quality of a learned grammar. Indeed, a grammar is equivalent to a binary classification system labeled with $\{1, 0\}$. For grammars we can assume that the sequences belonging to the grammar are data labeled with $1$, and non-grammar sequences are data labeled with $0$. This metric is therefore able to test to what extent the learned domain $D$ can regenerate the grammar. A domain $D$ can regenerate a grammar if $D$ accept, i.e. can generate, all positive test sequences $e \in E^{+}$ and reject, i.e. cannot generate, test negative sequences $e \in E^{-}$. Formally, the FScore is computed as follows: $ \mbox{FScore} = \frac{2.P.R}{P+R}$ where $R$ is the recall, i.e. the rate of sequences $e$ accepted by the ground truth domain that are successfully accepted by the learned domain, computed as follows: $R = \frac{|\{e \in E^{+}~|~accept(D, e)\}|}{|E|}$ and $P$ is the precision, i.e. the rate of sequences $e$ accepted by the learned domain that are sequences accepted by the ground truth domain, computed as follow: $P = \frac{|\{e \in E^{+}~|~accept(D,e)\}|}{|\{e \in E^{+}~|~accept(D,e)\} \cup \{e \in E^{-}~|~accept(D,e)\}|}$.

\item {\em Accuracy:} It quantifies to what extent learned domains are able to solve new planning problems \cite{rim}. Most of the works addressing the problem of learning planning domains uses the syntactical error to quantify the performance of the learning algorithm. However, domains are learned to be used for planning, and it often happens that one missing precondition or effect, which amounts to a small syntactical error, makes them unable to solve planning problems. Formally, the accuracy $Acc = \frac{N}{N^{*}}$ is the ratio between $N$, the number of correctly solved problems with the learned domain, and $N^{*}$, the total number of problems to solve. The accuracy is computed over 20 problems. We also report in our results the ratio of (possibly incorrectly) solved problems. In the experiments, we solve the test problems with different planners. Instances of Sequential domains and SHE domains are solved with the TP-SHE \cite{tp} planner and instances of the Cushing domain are solved with the Tempo planner \cite{tp}. We use different planners because some planners have good results with only some forms of action concurrency. For instance, TP-SHE is the domain with the best performances for instances with Single Hard Envelope but has bad results for Cushing. Plan validation is realized with the automatic validation tool used in the IPC competition VAL \cite{val}.

\item {\em IPC quality score:} This metric is initially used to compared different planners. We can use to test a learned domain. The score of a solved problem is the ratio $C^{*}/C$ where $C$ is the cost of the plan found by the learned domain and $C^{*}$ is the cost of the plan founded by the reference domain. The cost of a plan is the sum of all durations of all actions of the plan. The score on an unsolved problem is $0$. The score of a learned domain is the sum of its scores for all problems.
\end{itemize}

\subsection{Results}

\begin{table*}[!t]
\centering
\begin{subtable}[h]{\textwidth}
\resizebox{\textwidth}{!}{
\begin{tabular}{|c|c|c|c|c|c|c|c|c|}
\hline
Domain & Algorithm & $E_\sigma$& FScore (classical)& FScore (temp) & $Solved$& $Acc$& $IPC$& Time (sec) \\
\hline
\multirow{2}{*}{Peg}
& TempAMLSI & $2.8 \%$ & $100 \%$ & $100 \%$ & $100 \%$ & $100 \%$ & $20$ & $4.2$ \\
& TempAMLSI+TR & $2.8 \%$ & $100 \%$ & $100 \%$ & $100 \%$ & $100 \%$ & $20$ & $3.2$ \\ \hline
\multirow{2}{*}{Sokoban}
& TempAMLSI & $0.1 \%$ & $100 \%$ & $100 \%$ & $100 \%$ & $100 \%$ & $19.9$ & $23.4$ \\
& TempAMLSI+TR & $0.1 \%$ & $100 \%$ & $100 \%$ & $100 \%$ & $100 \%$ & $19.9$ & $27$ \\ \hline
\multirow{2}{*}{Parking}
& TempAMLSI & $5.6 \%$ & $100 \%$ & $100 \%$ & $100 \%$ & $100 \%$ & $19.9$ & $62.3$ \\
& TempAMLSI+TR & $5.6 \%$ & $100 \%$ & $100 \%$ & $100 \%$ & $100 \%$ & $19.9$ & $74.2$ \\ \hline
\multirow{2}{*}{Zenotravel}
& TempAMLSI & $0.7 \%$ & $72.1 \%$ & $100 \%$ & $100 \%$ & $100 \%$ & $20$ & $20.1$ \\
& TempAMLSI+TR & $0.7 \%$ & $72.1 \%$ & $100 \%$ & $100 \%$ & $100 \%$ & $20$ & $20$ \\ \hline
\end{tabular}
}
\caption{\label{tab:results:sequential} Domain learning results on sequential temporal domains.}
\end{subtable}
\begin{subtable}[h]{\textwidth}
\resizebox{\textwidth}{!}{
\begin{tabular}{|c|c|c|c|c|c|c|c|c|}
\hline
Domain & Algorithm & $E_\sigma$& FScore (classical)& FScore (temp) & $Solved$& $Acc$& $IPC$& Time (sec) \\
\hline
\multirow{2}{*}{Turn and Open}
& TempAMLSI & $3.2 \%$ & $82.4 \%$ & $100 \%$ & $55 \%$ & $55 \%$ & $10.7$ & $36.6$ \\
& TempAMLSI+TR & $3.2 \%$ & $82.4 \%$ & $100 \%$ & $55 \%$ & $55 \%$ & $10.7$ & $37.3$ \\ \hline
\multirow{2}{*}{Match}
& TempAMLSI & $5.3 \%$ & $86 \%$ & $82.3 \%$ & $93.3 \%$ & $93.3 \%$ & $18.7$ & $1.8$ \\
& TempAMLSI+TR & $2.7 \%$ & $87.4 \%$ & $100 \%$ & $93.3 \%$ & $93.3 \%$ & $18.7$ & $4.3$ \\ \hline
\end{tabular}
}
\caption{\label{tab:results:she} Domain learning results on SHE temporal domains.}
\end{subtable}
\begin{subtable}[h]{\textwidth}
\resizebox{\textwidth}{!}{
\begin{tabular}{|c|c|c|c|c|c|c|c|c|}
\hline
Domain & Algorithm & $E_\sigma$& FScore (classical)& FScore (temp) & $Solved$& $Acc$& $IPC$& Time (sec) \\
\hline
\multirow{2}{*}{Cushing}
& TempAMLSI & $7.9 \%$ & $83.6 \%$ & $47.4 \%$ & $0 \%$ & $0 \%$ & $0$ & $23.9$ \\
& TempAMLSI+TR & $8.2 \%$ & $90.7 \%$ & $90.6 \%$ & $0 \%$ & $0 \%$ & $0$ & $451.3$ \\ \hline
\end{tabular}
}
\caption{\label{tab:results:cushing} Domain learning results on Cushing temporal domain.}
\end{subtable}
\caption{\label{tab:results} Domain learning results on 7 domains when observations are complete. TempAMLSI performance is measured in terms of, syntactical error $E_\sigma$, FScore for temporal and classical samples, accuracy $Acc$, IPC Score and runtimes.}
\end{table*}

\begin{figure*}[!t]
\centering
    \begin{subfigure}[b]{0.3\textwidth}
        \centering
        \includegraphics[width=.99\textwidth]{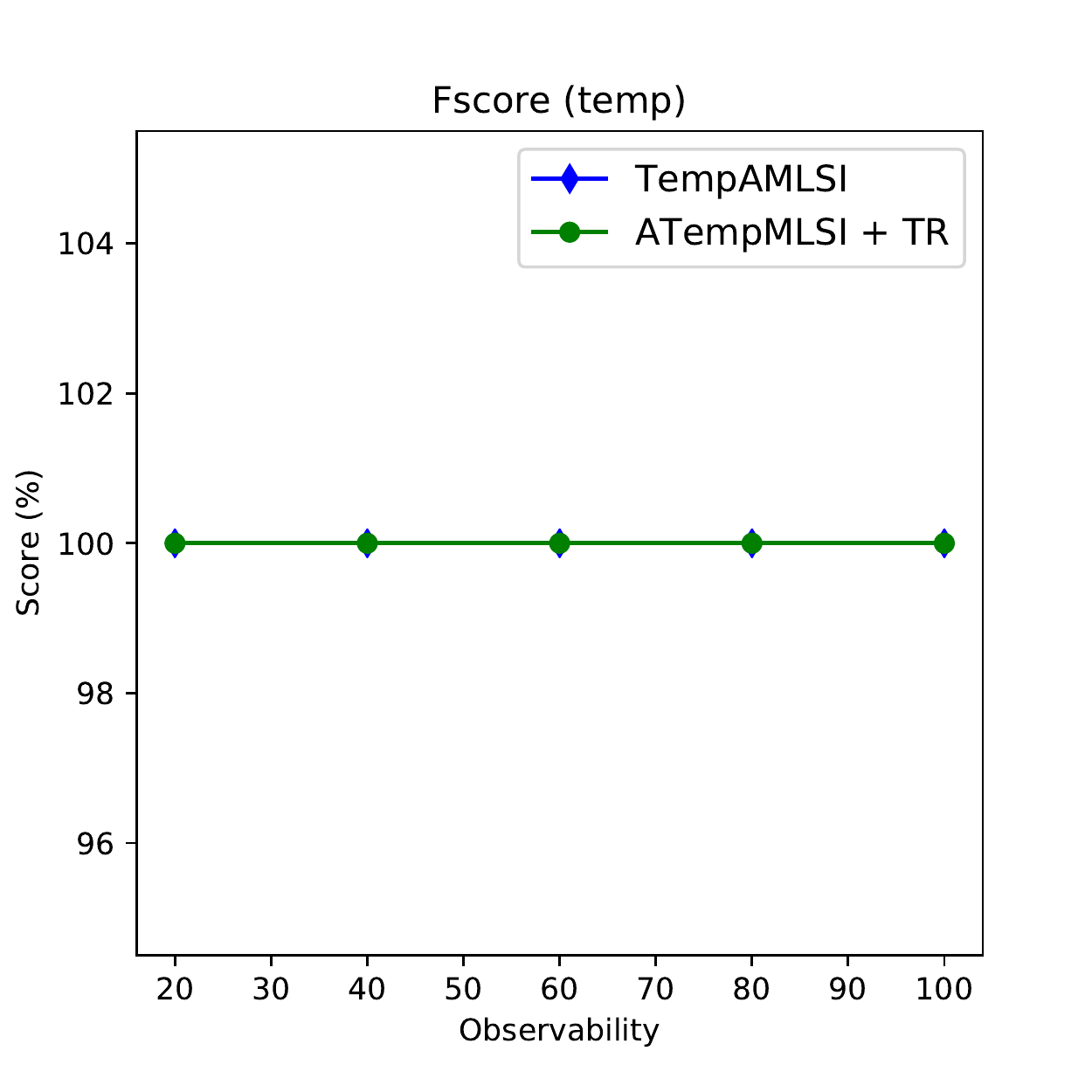}
        \caption{Peg Solitaire}
        \label{fig:peg}
    \end{subfigure}
    \begin{subfigure}[b]{0.3\textwidth}
        \centering
        \includegraphics[width=.99\textwidth]{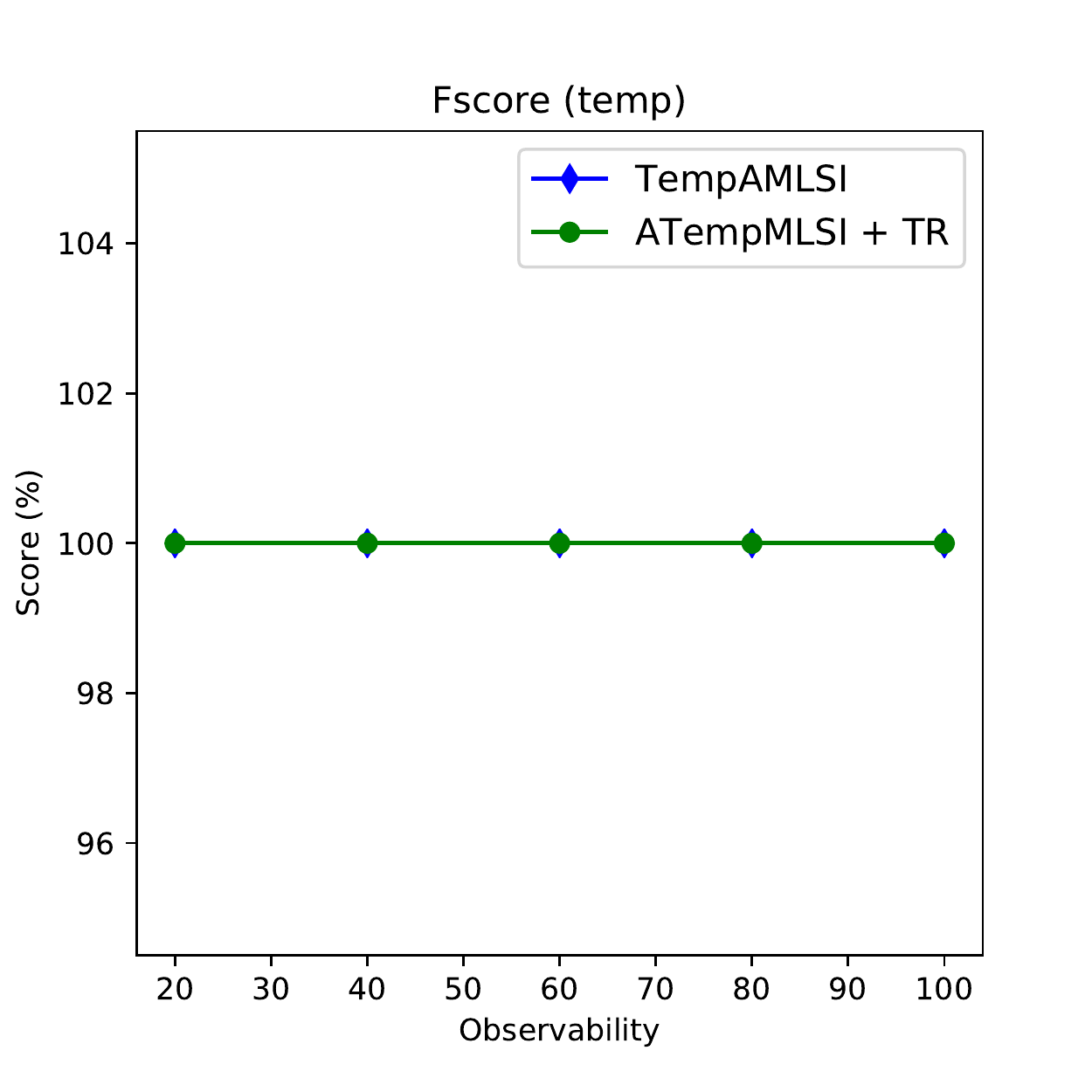}
        \caption{Sokoban}
        \label{fig:sokoban}
    \end{subfigure}
    \begin{subfigure}[b]{0.3\textwidth}
        \centering
        \includegraphics[width=.99\textwidth]{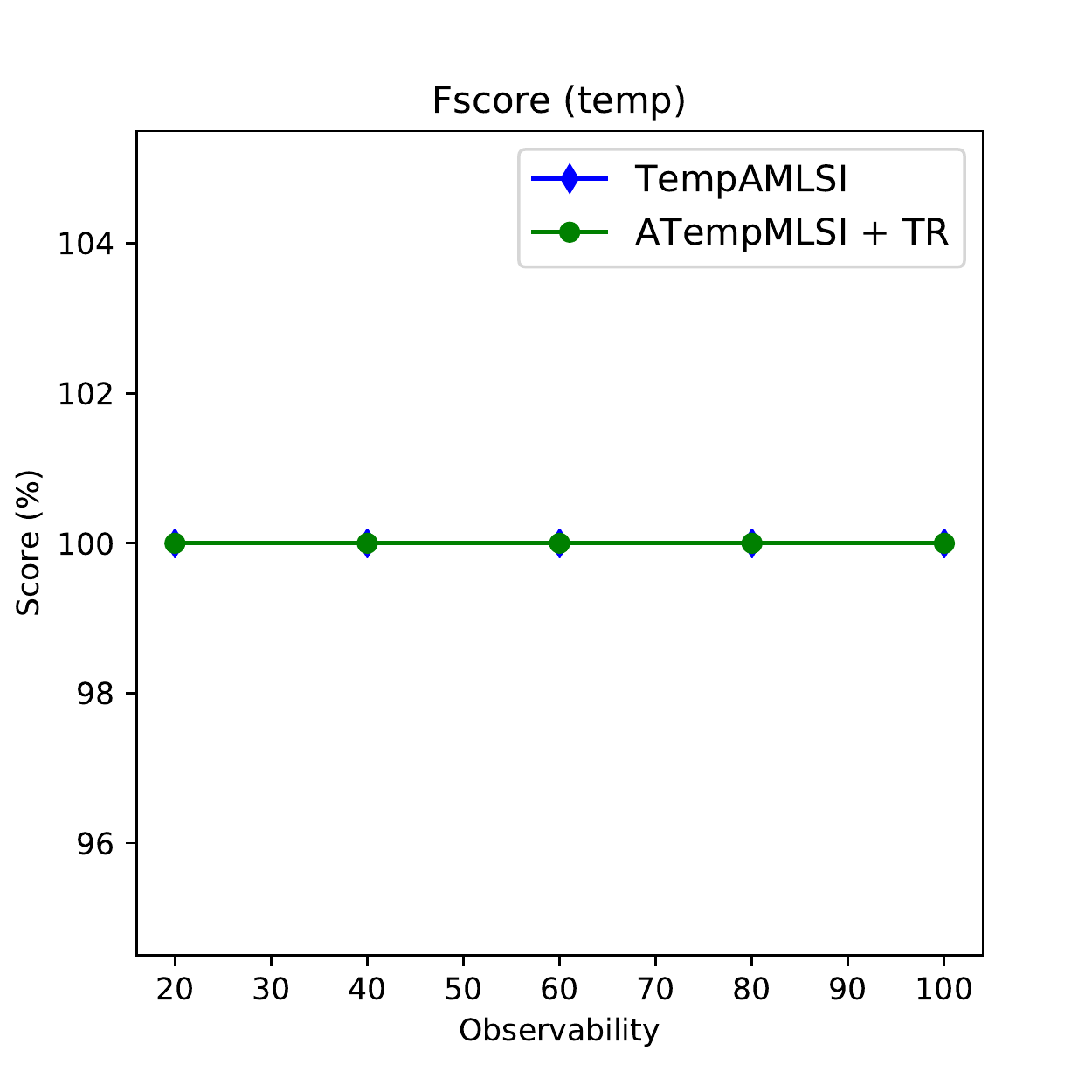}
        \caption{Zenotravel}
        \label{fig:zenotravel}
    \end{subfigure}
     \begin{subfigure}[b]{0.3\textwidth}
        \centering
        \includegraphics[width=.99\textwidth]{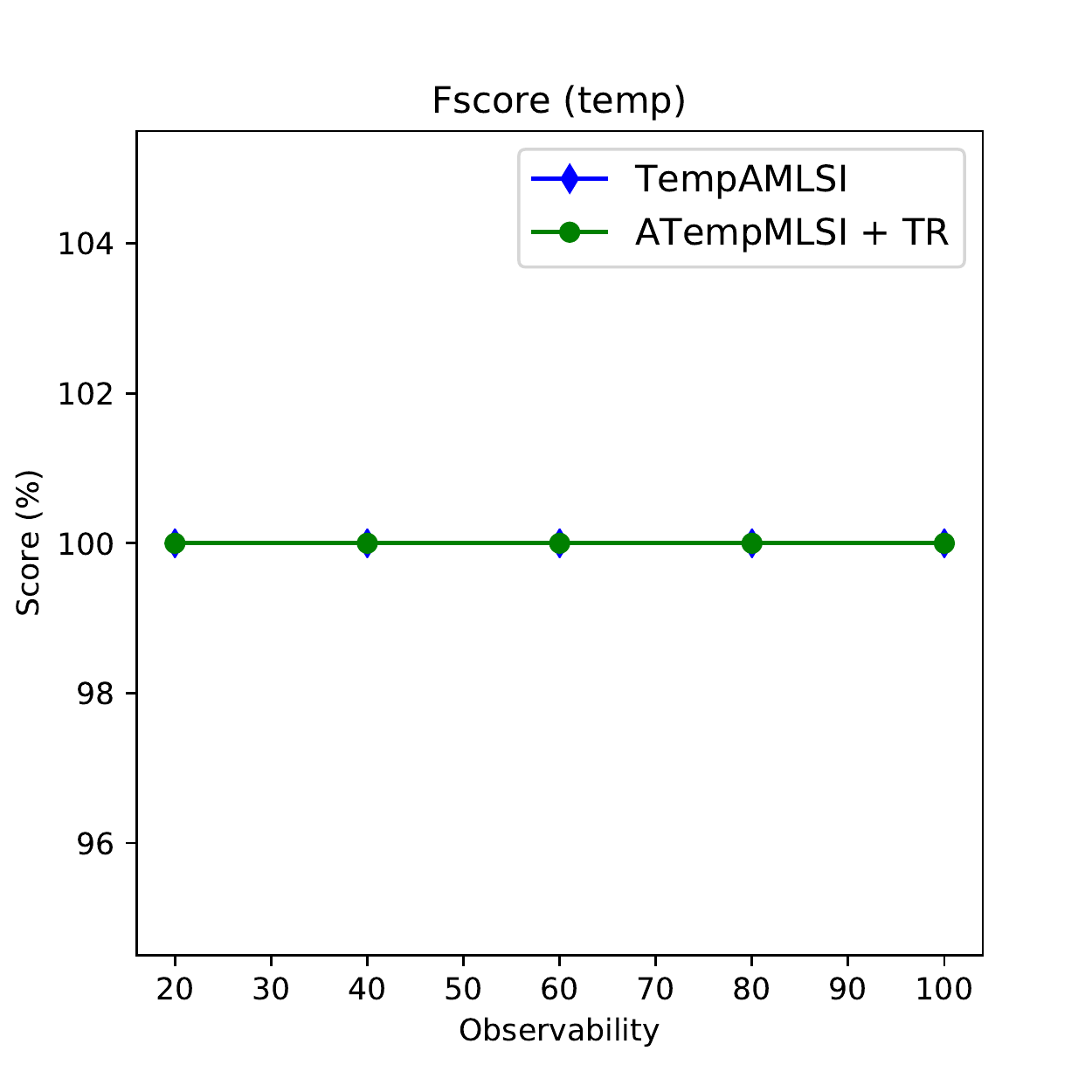}
        \caption{Parking}
        \label{fig:parking}
    \end{subfigure}
    \caption{Learned domains Syntactical distance with different levels of observability}
    \label{fig:experiment_partial_seq}
\end{figure*}

\begin{figure*}[!t]
\centering
    \begin{subfigure}[b]{0.3\textwidth}
        \centering
        \includegraphics[width=.99\textwidth]{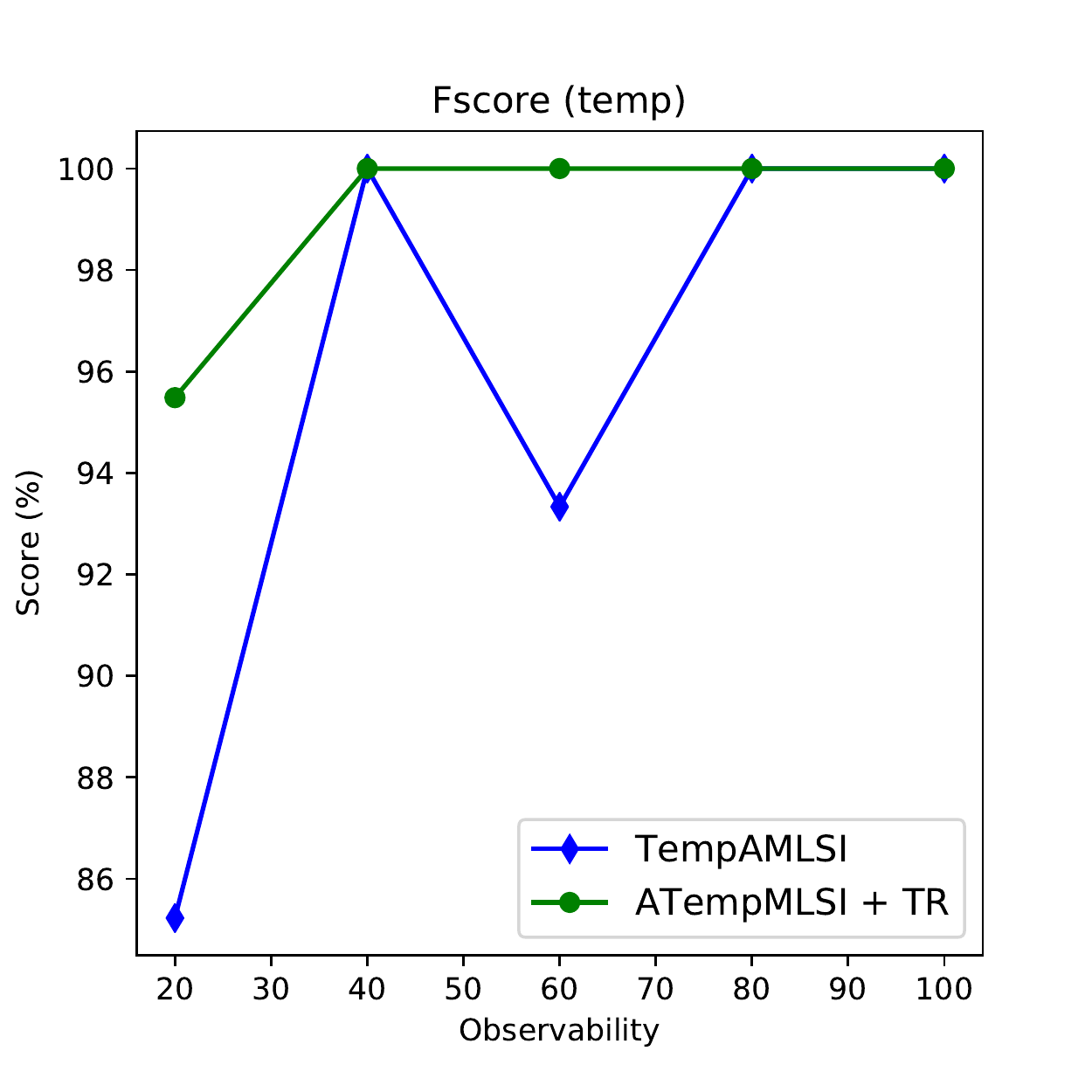}
        \caption{Turn and Open}
        \label{fig:turn-and-open}
    \end{subfigure}
    \begin{subfigure}[b]{0.3\textwidth}
        \centering
        \includegraphics[width=.99\textwidth]{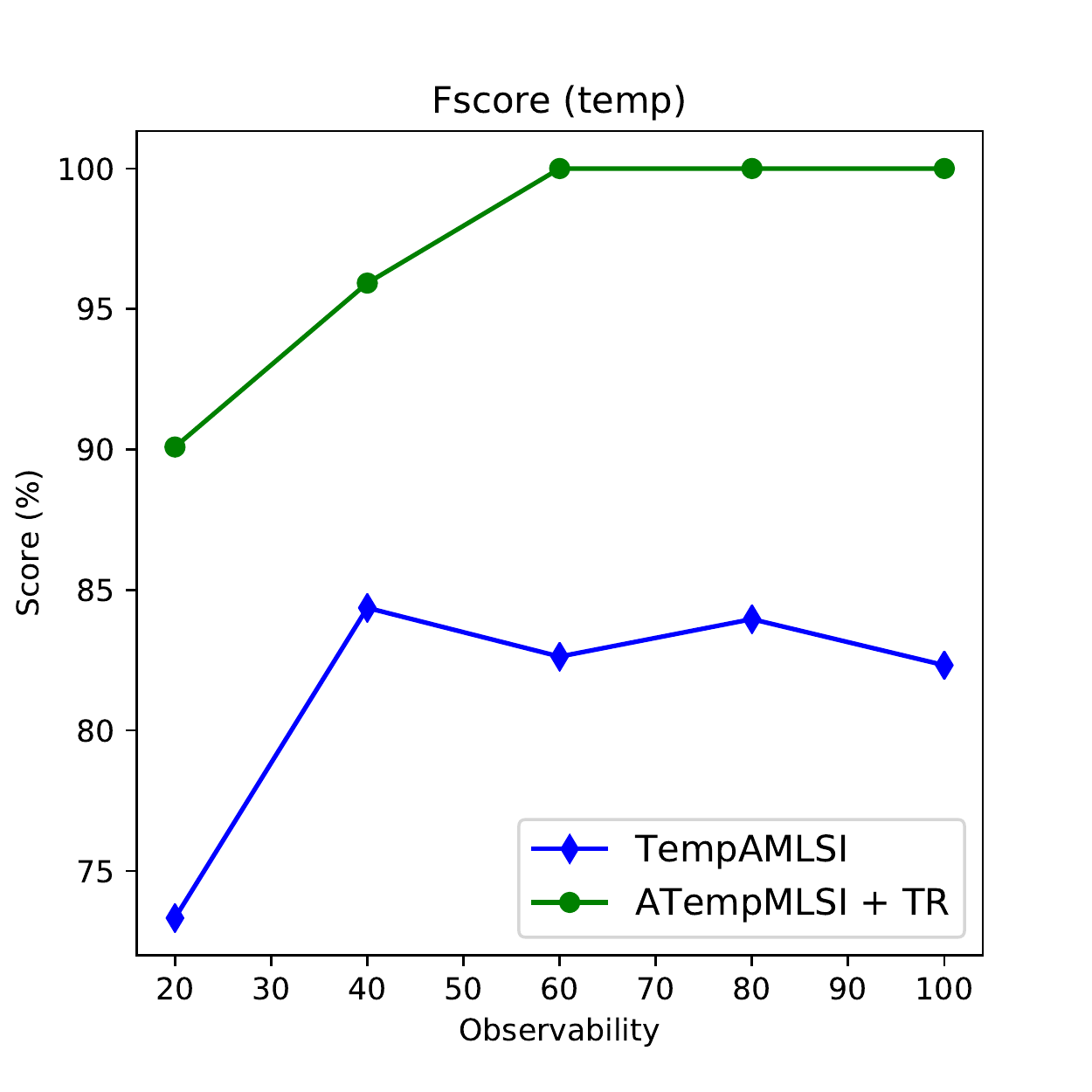}
        \caption{Match}
        \label{fig:match}
    \end{subfigure}
    \caption{Learned domains Syntactical distance with different levels of observability}
    \label{fig:experiment_partial_she}
\end{figure*}

\begin{figure*}[!t]
\centering
    \begin{subfigure}[b]{0.3\textwidth}
        \centering
        \includegraphics[width=.99\textwidth]{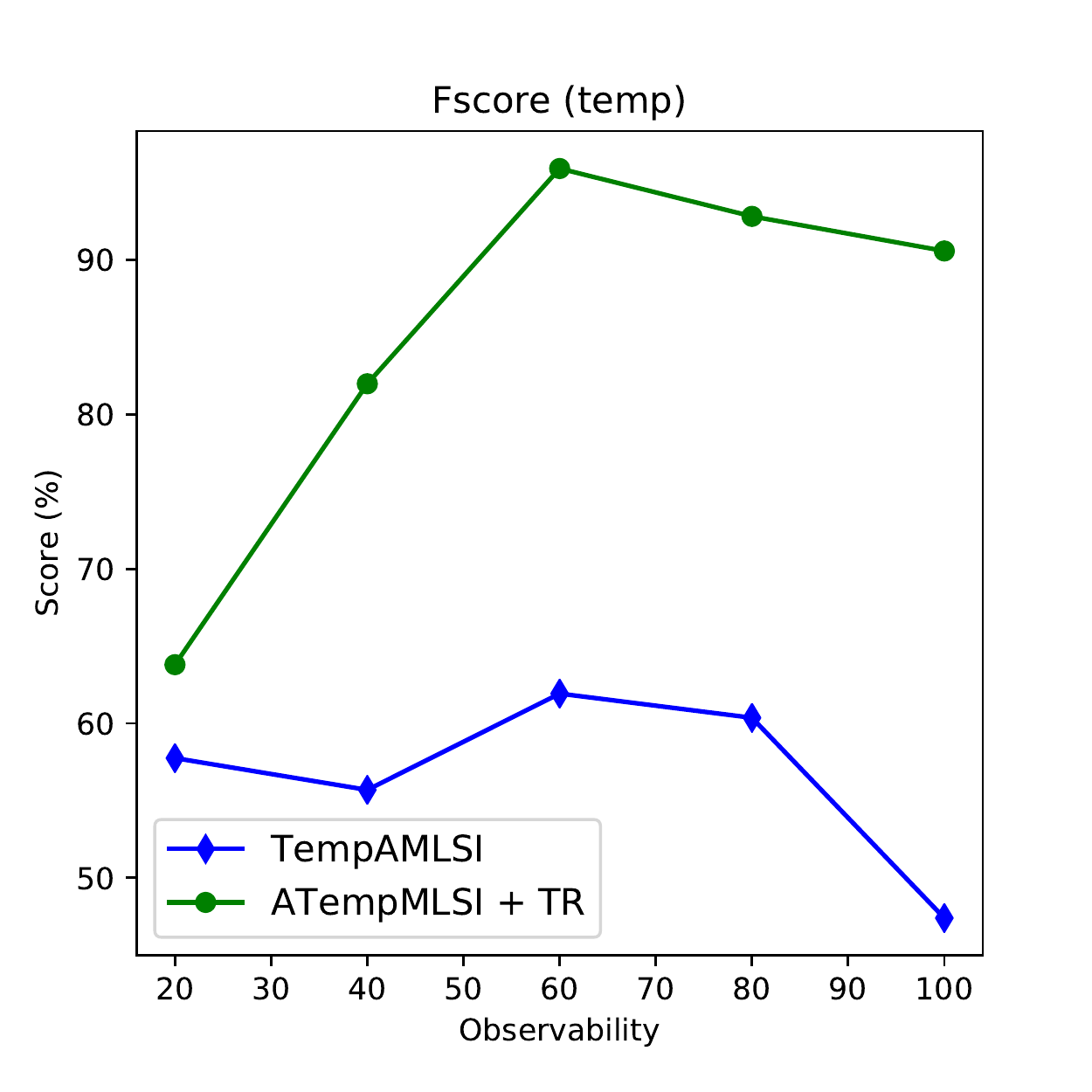}
        \caption{Cushing}
        \label{fig:cushing}
    \end{subfigure}
    \caption{Learned domains Syntactical distance with different levels of observability}
    \label{fig:experiment_partial_cushing}
\end{figure*}

In this section we perform an ablation study and compare two variants: (1) \textit{TempAMLSI:} TempAMLSI only uses the AMLSI algorithm to learn temporal domain and (2) \textit{TempAMLSI + TR:} TempAMLSI uses both AMLSI algorithm and temporal refinement to learn temporal domains.

\subsubsection{Learning with complete observations}

Table \-- \ref{tab:results} shows TempAMLSI's performance when observations are complete.

First of all, Table \-- \ref{tab:results:sequential} shows results for Sequential domains. Firstly, we can note that for each domains TempAMLSI and TempAMLSI+TR variants have same performances, whatever the metric. Then, we observe that learning runtimes is generally higher for TempAMLSI+TR variants. For the Peg Solitaire domain we can observe that learned domains are accurate. More precisely we observe that both classical and temporal $FScore$ and $accuracy$ are optimal, also we observe that $IPC = 20$ that implies that plans generated with learned domains are identical to the plans generated with the ground truth domain. Also, the syntactical distance is not optimal ($E_\sigma = 2.8\%$), this is due to the fact that some implicit preconditions which are not encoded in the reference domain and encoded in the learned domain. Then, for the Sokoban and Parking domain results are similar. However, we can note that the IPC score is not optimal ($IPC = 19.9$ for Sokoban and Parking), this is due to the fact that some plans found by the learned domains has a higher cost than the plan found by the reference domain. Then, for the Zenotravel domain we can observe that Temporal Fscore and accuracy are optimal. Also, we observe that the classical FScore is not optimal, this is due to the fact that some preconditions encoded as {\em at start} preconditions in the reference domain are encoded as {\em overall} preconditions in the learned domains.

Then, Table \-- \ref{tab:results:she} gives results for SHE domains. For the Turn and Open domain we can observe that, whatever the variant used, only the temporal FScore is optimal. This is due to the fact that the Turn and Open domain contains SHE. Also we can note that the learned domains are able to solve the majority of new problems ($Acc = 55\%$). For this domain, we observe that the temporal refinement has no impact, this is due to the fact that the Turn and Open domain contains both sequential operators ($move$, $pick$, $drop$) and SHE operators ($open-door$, $turn-doorknob$). Then, for the Match domain we can observe that the domain learned with temporal refinement has better results for syntactical distance and both classical and temporal FScore. A better temporal Score means that the domains learned with the TempAMLSI+TR variant better respect the forms of required action concurrency of the ground truth domain. However, TempAMLSI and TempAMLSI+TR variants have the same accuracy ($Acc = 93.3\%$). Finally, we observe that learning runtimes are higher for the TempAMLSI+TR variant for all SHE domains.

Finally, for the Cushing domain (see Table \-- \ref{tab:results:cushing}) we observe that the temporal refinement step only increases the temporal FScore. This implies that the forms of concurrency present in the domains learned with the TempAMLSI+TR variant are closer to the forms of concurrency in the ground truth domain than the domains learned with the TempAMLSI variant. Also, the TempAMLSI variant gives better syntactical distance and classical FScore than the TempAMLSI+TR variant. Also, we observe that the learned domains cannot solve problem whatever the variant used. Finally, we note that the temporal refinement strongly increases the learning runtimes.

\subsubsection{Learning with partial observations}

Figures \-- \ref{fig:experiment_partial_seq}, \ref{fig:experiment_partial_she} and \ref{fig:experiment_partial_cushing} show how temporal FScore varies when the level of observed proposition in intermediate states varies.

First, we can observe that for all sequential domains (see Figure \-- \ref{fig:experiment_partial_seq}) the temporal refinement step has no effect. Also, Temporal FScore is always optimal whatever the level of observability. Then, Table \-- \ref{fig:experiment_partial_she} shows that, for SHE domains, the TempAMLSI+TR variant gives better results. We observe that FScore is optimal when at least $40 \%$ of propositions are observed for the Turn and Open and when at least $60 \%$ of propositions are observed for the Match domain. Finally, Table \-- \ref{fig:experiment_partial_cushing} shows that for the Cushing domain the TempAMLSI+TR variant gives better. However, TempAMLSI+TR never gives the optimal FScore.

\section{Conclusion}

In this paper we presented TempAMLSI, a novel algorithm to learn temporal PDDL domains. TempAMLSI is based on the AMLSI approach. In this paper we reused the idea to use classical PDDL domain proposed by several temporal planners. More precisely, TempAMLSI converts the temporal sample into a sample containing non-temporal sequences, then TempAMLSI uses the AMLSI algorithm to learn a classical PDDL domain and convert it into a temporal PDDL domain, Also, TempAMLSI has a temporal refinement step allowing to deal with different forms of required action concurrency. Finally, we show experimentally that the TempAMLSI approach was able to learn accurate domains with sequential sequences and single hard envelopes. In future works, TempAMLSI will be extended to learn temporal PDDL domain with noisy observations and temporal PDDL domain with other form of required action concurrency than Single Hard Envelopes. Also, TempAMLSI will be extended to be able to deal with different ways for the durative action conversion, such as the LGP translation \cite{lpgp} for instance.

\section*{Acknowledgments}
This research is supported by the French National Research Agency under the "Investissements d’avenir” program (ANR-15-IDEX-02) on behalf of the Cross Disciplinary Program CIRCULAR.

\bibliography{biblio}
\bibliographystyle{aaai}
\end{document}